\documentclass[letterpaper, 10 pt, conference]{ieeeconf}
\IEEEoverridecommandlockouts

\usepackage{lipsum}
\usepackage{graphicx}
\usepackage{cite}
\usepackage{xcolor}
\usepackage{amsmath} 
\usepackage{amssymb}
\usepackage{algorithm}
\usepackage{algorithmic}
\definecolor{targetarea}{HTML}{8FB88F}
\begin{document}

\title{Safety Control of a Hyper-redundant Robot via Adaptive Weighted Control Barrier Functions}

\author{Zijian Cai, 
Kiwan Wong, 
Wenci Xin, 
Wei Xiao, 
Daniela Rus, 
and Cecilia Laschi
\thanks{This work was funded by the National Research Foundation (NRF), Prime Minister’s Office, Singapore under its Campus for Research Excellence and Technological Enterprise (CREATE) programme. The Mens, Manus, and Machina (M3S) is an interdisciplinary research group (IRG) of the Singapore MIT Alliance for Research and Technology (SMART) centre \textit{(Z. Cai and K. Wong contributed equally to this work. Corresponding: wenci.xin@smart.mit.edu; wei.xiao@ntu.edu.sg)} }
\thanks{Z. Cai and W. Xin are with the Singapore-MIT Alliance for Research and Technology (SMART) Centre, Singapore}
\thanks{K. Wong and D. Rus are with the CSAIL, Massachusetts Institute of Technology, Cambridge, USA}
\thanks{W. Xiao is with the School of Electrical and Electronic Engineering, Nanyang Technological University, Singapore}
\thanks{C. Laschi is with the Department of Mechanical Engineering and Advanced Robotics Centre, National University of Singapore, Singapore}
\thanks{W. Xiao, D. Rus and C. Laschi are PIs of the Singapore MIT Alliance for Research and Technology (SMART) centre, Singapore}
}

\maketitle

\begin{abstract}
Hyper-redundant robots are well suited for confined-space manipulation due to their high dexterity, but safe operation in cluttered environments remains challenging. In addition, their slender structures often lead to uneven load distributions and nonuniform tracking errors along the body. To address these issues, this work proposes a weighted control barrier functions (W-CBFs) framework that enforces safety constraints while reducing tracking errors caused by uneven loading. The proposed controller was first evaluated on a circular path-following task under different obstacle configurations. With fixed weights, compared to the non-weighted method, the maximum reduction in root-mean-square (RMS) tracking error was 59.6\% in simulation and 87.7\% in physical experiments. An adaptive weighting strategy was then investigated based on the discrepancy between simulated and experimental performance under different mapping functions. The RMS errors were further reduced by 21.9\% and 8.5\%, respectively, although the error increases when obstacles were located close to the robot body. Finally, the robot was evaluated in a cleaning task requiring coverage of a rectangular area and compared with manual teleoperation. Although the controller was not explicitly optimized for area coverage, the autonomous strategy achieved comparable or better coverage performance while avoiding collisions with the surrounding frame, whereas collisions occurred during manual operation.
\end{abstract}

\IEEEpeerreviewmaketitle

\section{Introduction}
Thanks to their large number of degrees of freedom (DoFs), hyper-redundant robots can perform highly dexterous motions, making them well suited for tasks in deep and confined environments, such as industrial inspection \cite{lakhal2019control,chen2024review} and manipulation in restricted spaces \cite{mu2022hyper,liu2022review}. Among these applications, cleaning in cluttered environments, including lecture halls and operating theaters, presents a particularly challenging scenario, especially when surrounding objects are fragile and must be carefully avoided during operation. Existing cleaning robots are typically designed for relatively open environments and therefore have limited capability to access narrow gaps or clean hard-to-reach areas without interfering with their surroundings \cite{megalingam2025cleaning}. In these cluttered environments, only limited free space is available for robot motion, requiring the robot to combine sufficient reachability and extensibility with a high degree of dexterity.

\begin{figure}[!t]
    \centering
    \includegraphics[width=0.45\textwidth]{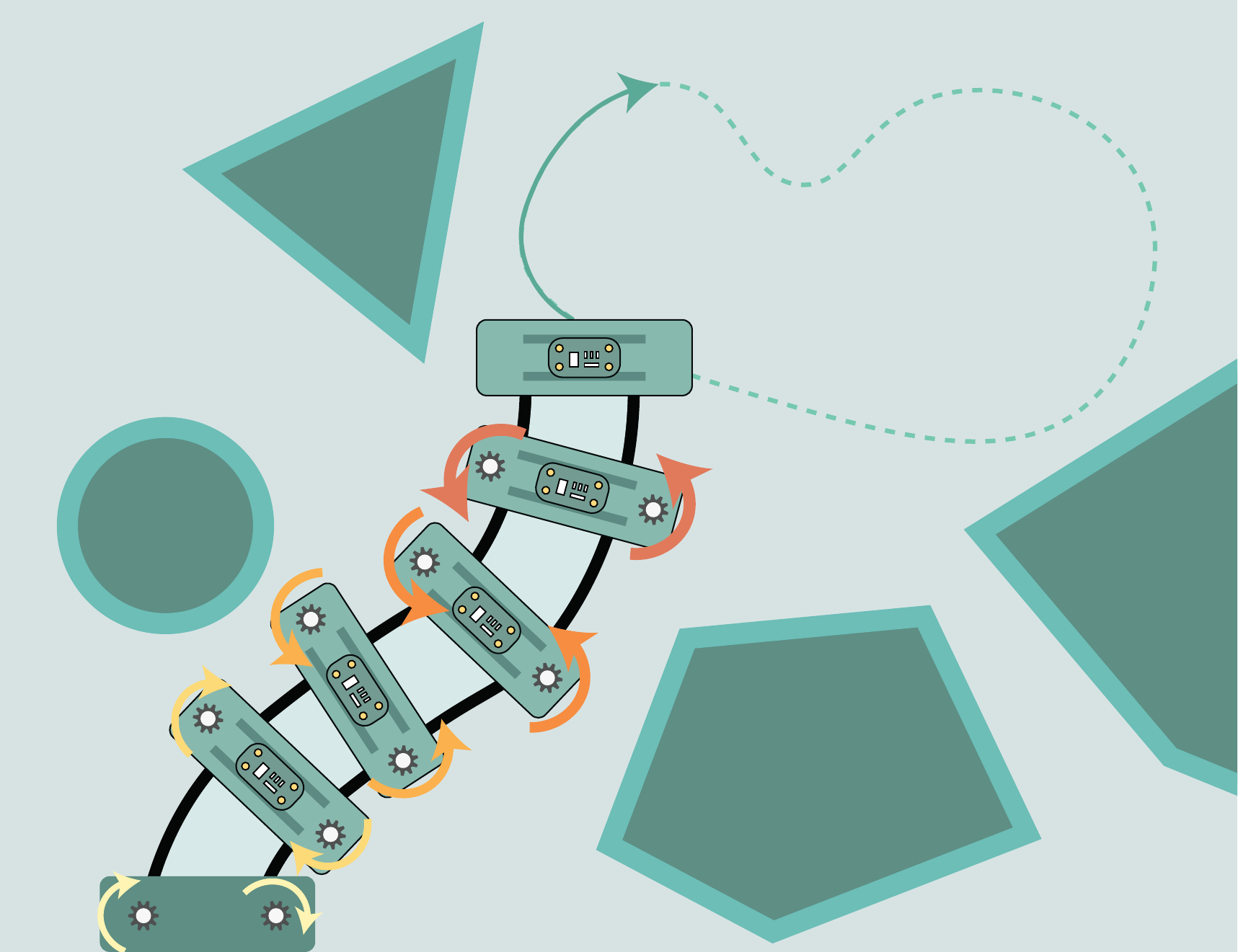}
    \caption{Schematic of the hyper-redundant robot safety control with weighted control barrier functions (W-CBFs).}
    \label{fig1}
    \vspace{-0.2in}
\end{figure}

With the development of soft robotics, various types of extensible hyper-redundant robots have been developed, employing mechanisms such as unwinding-based extension \cite{palmer2019active,yuan2025design}, pre-compression \cite{gong2025novel,poka2025alternative}, and rod-driven extension \cite{wang2022design}. In particular, Song et al. developed a planar extensible robot using flexible racks to enable the independent extension of multiple robot modules, demonstrating its potential for cleaning applications \cite{song2026shape}. Despite these advances in mechanical design, effectively controlling the large number of DoFs remains a significant challenge, particularly in determining how the redundant DoFs should be coordinated to accomplish a given task. Accordingly, various control strategies have been proposed. A basic approach is redundancy resolution through null-space projection, in which the redundant DoFs are exploited to achieve secondary objectives without affecting the primary end-effector task \cite{dietrich2015overview,pistone2024modelling}. More constrained strategies regulate the motion of the robot body by prescribing its overall shape or trajectory. Representative examples include backbone-curve-based methods, which describe the robot configuration using a continuous body curve \cite{bordini2025shape,peng2021end}, and follow-the-leader approaches, in which the robot body follows the trajectory traced by its tip when navigating through narrow spaces \cite{luo2024local,russo2023continuum}.

Despite the great advances in motion planning, safety is often treated via trajectory design and geometric constraints. The approach becomes problematic when the robot needs to operate in cluttered environments, where not only the end-effector but the whole body needs to remain collision-free. A trajectory that is safe for the tip might cause intermediate contact in between the rest of the body with respect to the environment. Therefore, safe control for hyper-redundant robots requires an algorithm to enforce whole-body safety continuously while progressing.

Control Barrier Functions (CBFs) \cite{Ames2017, xiao2021high} provide a natural framework for imposing such safety requirements into feedback control. By expressing collision avoidance and other physical limits, CBF-based controllers can modify a nominal command when necessary to maintain safety. The runtime filtering structure is particularly attractive for hyper-redundant robots. However, conventional formulations typically treat control correction under uniform control metric, implicitly treating actuation across different sections as equally preferable. 


Accordingly, our contributions are
\begin{itemize}
\item \textbf{Whole-Body Safety:} We formulate whole-body safety control for an extensible hyper-redundant robot, enforcing collision-avoidance constraints along the robot body rather than only at the end effector.

\item \textbf{Adaptive Weighting:} We propose an adaptive weighting scheme driven by measured kinematic discrepancy, which penalizes control inputs associated with less reliable robot sections. Hardware experiments further reveal that a uniform control metric can perform poorly due to accumulated friction and deformation in proximal sections.

\item \textbf{Real-time deployment:} We employ closed-form cyclic projections to realize computationally efficient safety filtering, enabling whole-body safety control at 25 Hz on a Raspberry Pi for a 36-body robotic system.

\end{itemize}

\section{Robotic Platform}
\begin{figure}[!t]
    \centering
    \includegraphics[width=0.5\textwidth]{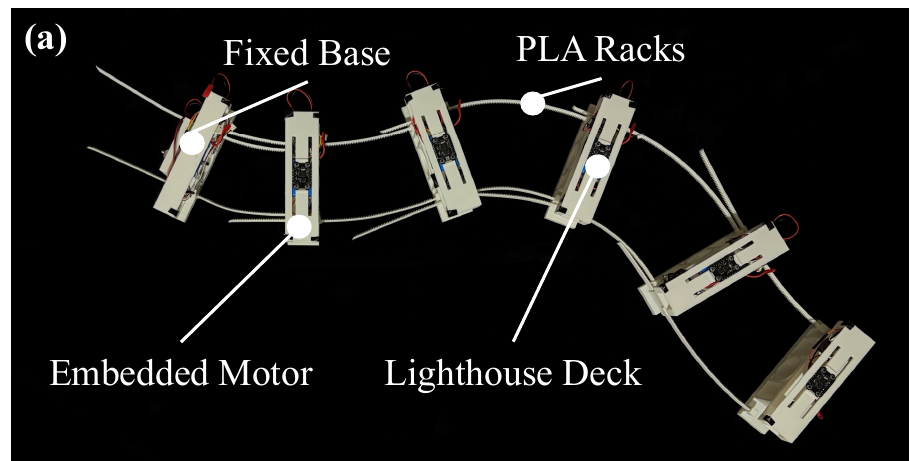}\\
    \includegraphics[width=0.5\textwidth]{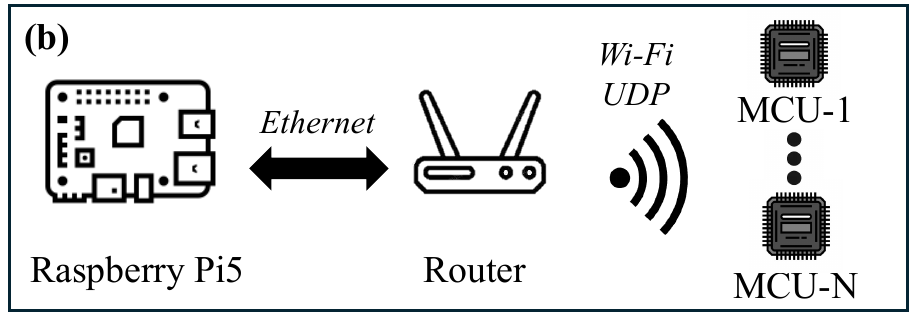}
    \caption{(a) An illustration of the hyper-redundant robot system including structure, actuation, and positioning. (b) The wireless communication framework between the Raspberry Pi and the MCUs.}
    \label{fig2}
    \vspace{-0.2in}
\end{figure}

Similar to Song et al. \cite{song2026shape}, we employ a rack-based 5-segment hyper-redundant robot. As shown in Fig. \ref{fig2}a, it consists of a fixed base and five movable rigid frames which are connected by flexible rack pairs. Two racks in each frame are independently actuated by motors, providing ten actuation inputs. The common extension of the two racks changes the length of the segment, while differential lengths produce bending. Coordinating these motions across the five sections enables the robot to extend and bend within the plane. To avoid interference, the distance between two racks in each segment widens from 92~mm at the base to 156~mm at the tip, resulting in the same rack differential producing a smaller bend angle toward the tip.

Although the displacement of each rack can be measured using the motor-embedded encoders, deformation of the flexible racks may lead to discrepancies between the actual robot configuration and the kinematic prediction derived from the encoder readings. To obtain direct measurements of the robot body configuration, we employed a Lighthouse positioning system consisting of two base stations and a tracking deck mounted on each movable frame \cite{taffanel2021lighthouse}.

In order to reduce wiring issues and improve modularity, the system is set up as a wireless version. As Fig. \ref{fig2}b shows, we used a Raspberry Pi 5 as the central computer, running the ROS2 control and positioning nodes. One ESP32-S3-Tiny Microcontroller (MCU) was implemented in each segment to interface with two motors and communicate with the central computer. Commands and sensor feedback are exchanged over Wi-Fi using User Datagram Protocol (UDP). 

\section{Weighted CBF Controller for the Hyper-Redundant Robot}

\subsection{Robot Kinematics}

We model the planar hyper-redundant robot using a piecewise-constant-curvature (PCC) representation. The robot consists of \(N\) serial sections, each actuated by two flexible racks. Let \(q_{i,L}\) and \(q_{i,R}\) denote the left and right rack lengths of section \(i\). Its arc length \(L_i\) and bending angle \(\theta_i\) are
\begin{equation}
L_i=\frac{q_{i,L}+q_{i,R}}{2},
\qquad
\theta_i=\frac{q_{i,L}-q_{i,R}}{c_i},
\label{eq:pcc_coordinates}
\end{equation}
where \(c_i\) is the rack separation. Equal rack motion produces extension, whereas differential motion produces bending. The pose of each section boundary and points along the robot backbone are obtained by sequentially composing the PCC transformations.

We define the rack-length state and rack-velocity control input as
\begin{equation}
x=
[q_{1,L},q_{1,R},\ldots,q_{N,L},q_{N,R}]^\top,
\end{equation}
\begin{equation}
u=
[\dot q_{1,L},\dot q_{1,R},\ldots,\dot q_{N,L},\dot q_{N,R}]^\top.
\end{equation}
Since the actuators are operated in velocity control, the kinematic system takes the control-affine form
\begin{equation}
\dot{x}=f(x)+g(x)u,
\label{eq:control_affine}
\end{equation}
with
\begin{equation}
f(x)=0,
\qquad
g(x)=I_{2N}.
\label{eq:fg_robot}
\end{equation}

Let \(p_j(x)\in\mathbb{R}^2\) denote the Cartesian position of the \(j\)-th point along the robot body.
For a backbone point at normalized position $s_j\in[0,1]$
along section $i$, its Cartesian position is
\begin{equation}
p_j(x)=o_{i-1}
+R(\psi_{i-1})\frac{L_i}{\theta_i}
\begin{bmatrix}
\sin(s_j\theta_i)\\
\cos(s_j\theta_i)-1
\end{bmatrix},
\end{equation}
where $R$ is the planar rotation matrix, and
$o_{i-1}$ and $\psi_{i-1}$ are the section's starting
position and heading obtained recursively from preceding
sections, with $\psi_i=\psi_{i-1}-\theta_i$.
At $\theta_i=0$, the local displacement is
$[s_jL_i,0]^\top$.
Its velocity is given by
\begin{equation}
\dot p_j
=
J_j(x)\dot x
=
J_j(x)\bigl(f(x)+g(x)u\bigr),
\label{eq:point_velocity}
\end{equation}
where
\begin{equation}
J_j(x)=\frac{\partial p_j(x)}{\partial x}
\end{equation}
is the Jacobian of the \(j\)-th body point with respect to the robot state. This representation allows task objectives and collision-avoidance constraints to be formulated in Cartesian space while retaining the rack velocities as the control inputs.

\subsection{Weighted Control Barrier Functions}

Let \(h(x)\) define the safe set
\begin{equation}
\mathcal C=\{x\mid h(x)\geq0\}.
\end{equation}
A control barrier function (CBF) enforces forward invariance of \(\mathcal C\) through
\begin{equation}
L_fh(x)+L_gh(x)u+\alpha(h(x))\geq0,
\label{eq:cbf_condition}
\end{equation}
where
\begin{equation}
L_fh=\nabla h^\top f,
\qquad
L_gh=\nabla h^\top g,
\end{equation}
and \(\alpha(\cdot)\) is an extended class-\(\mathcal K\) function. Equation~\eqref{eq:cbf_condition} can be written as the affine half-space
\begin{equation}
a^\top u\geq b,
\label{eq:cbf_halfspace}
\end{equation}
with
\begin{equation}
a=L_gh^\top,
\qquad
b=-L_fh-\alpha(h).
\end{equation}
For the kinematic model in \eqref{eq:fg_robot}, \(L_fh=0\) and \(L_gh=\nabla h^\top\).

A conventional Euclidean CBF correction treats all control directions equally. However, different sections of the physical robot can exhibit substantially different actuation accuracy due to kinematic discrepancy, i.e., accumulated friction, structural deformation, and load. We therefore introduce the positive-definite diagonal metric
\begin{equation}
W=
\operatorname{diag}
(w_1,w_1,\ldots,w_N,w_N),
\qquad
w_i>0,
\label{eq:weight_matrix}
\end{equation}
where both racks of section \(i\) share one weight.

Given a nominal command \(u_{\mathrm{nom}}\), the weighted CBF correction is
\begin{equation}
\begin{aligned}
u^\star
=
\arg\min_u \quad &
\frac{1}{2}
(u-u_{\mathrm{nom}})^\top
W
(u-u_{\mathrm{nom}}) \\
\mathrm{s.t.}\quad &
a^\top u \geq b .
\end{aligned}
\label{eq:wcbf_qp}
\end{equation}

For a single affine constraint, the KKT conditions yield the closed-form projection
\begin{equation}
\boxed{
u^\star
=
u_{\mathrm{nom}}
+
\frac{
[b-a^\top u_{\mathrm{nom}}]_+
}{
a^\top W^{-1}a
}
W^{-1}a
}
\label{eq:wcbf_closed_form}
\end{equation}
where \([z]_+=\max(0,z)\). If the nominal command already satisfies the CBF condition, no correction is applied. 

Equation~\eqref{eq:wcbf_closed_form} shows that the correction direction is \(W^{-1}a\), rather than \(a\) as in the Euclidean case. Therefore, increasing \(w_i\) reduces modification of the corresponding control inputs. The weighting changes only how the required correction is distributed among robot sections; it does not modify the CBF constraint or the safe set.

\subsection{Multiple Safety Constraints}

For multiple CBF constraints, define
\begin{equation}
\mathcal H_i
=
\{u\mid a_i^\top u\geq b_i\},
\qquad
\mathcal H
=
\bigcap_{i=1}^{M}\mathcal H_i.
\end{equation}
Rather than solving a multi-constraint QP, we sequentially apply the closed-form weighted projection to each half-space \cite{wong2026posafenet}. Let \(P_i^W\) denote the projection onto \(\mathcal H_i\). One cyclic sweep is
\begin{equation}
u^{k+1}
=
P_M^W
\cdots
P_2^W
P_1^W(u^k),
\qquad
u^0=u_{\mathrm{nom}}.
\label{eq:cyclic_projection}
\end{equation}

At each control step, $W$ is held fixed throughout the
projection iterations. Under $z=W^{1/2}u$, the weighted
projection of $v$ onto $a^\top u\geq b$ becomes
\begin{equation}
\min_z\ \frac12\|z-W^{1/2}v\|_2^2
\quad\mathrm{s.t.}\quad
(W^{-1/2}a)^\top z\geq b.
\end{equation}
Thus, weighted projections are equivalent to Euclidean
projections onto the transformed half-spaces.

Consequently, if the instantaneous half-spaces have a nonempty intersection, $\mathcal H\neq\emptyset$, repeated cyclic projections converge to a jointly feasible control \cite{bregman1965method}:
\begin{equation}
u^k
\rightarrow
u^\infty\in\mathcal H,
\qquad
k\rightarrow\infty.
\label{eq:projection_convergence}
\end{equation}
This guarantees convergence to feasibility for a fixed positive-definite \(W\), although the limiting point is not in general the exact weighted projection of \(u_{\mathrm{nom}}\) onto the full intersection.

A finite projection budget also does not guarantee exact joint feasibility, because a later projection can reintroduce violation of an earlier constraint. We therefore quantify the remaining violation using
\begin{equation}
r(u)
=
\max_{i=1,\ldots,M}
[b_i-a_i^\top u]_+.
\label{eq:projection_residual}
\end{equation}
Here, \(r(u)=0\) indicates joint feasibility, while a nonzero residual measures the maximum remaining CBF violation. The number of projection iterations therefore provides an explicit trade-off between computational cost and constraint satisfaction.

\subsection{Adaptive Weight Selection}

The fixed-weight controller requires selecting the relative importance of individual robot sections in advance. However, the accuracy of the nominal PCC model can vary with robot configuration and loading. We therefore adapt the weights using the discrepancy between the kinematic prediction and external pose measurements.

For section \(i\), let \(\Delta p_i\) and \(\Delta\theta_i\) denote the position and wrapped orientation residuals between adjacent frames, expressed in the preceding frame coordinates. Over an update window \(\mathcal B_k\), we define the model discrepancy of section \(i\) as
\begin{equation}
\eta_i[k]
=
\sqrt{
\frac{
\sum_{t\in\mathcal B_k}
\left(
\|\Delta p_i(t)\|_2^2+
[L_i(t)\Delta\theta_i(t)]^2
\right)
}{
\sum_{t\in\mathcal B_k}
\max(\bar L_i(t),L_0)^2
}
},
\label{eq:adaptive_indicator}
\end{equation}
where \(L_0>0\) prevents normalization by small extensions. The normalized discrepancy \(z_i\in[0,1]\)
\begin{equation}
z_i=\operatorname{clip}\!\left(\frac{\eta_i-\eta_{\min}}{\eta_{\max}-\eta_{\min}},0,1\right),
\label{eq:normalized_dis}
\end{equation}
is mapped to the section weight according to
\begin{equation}
w_i
=
w_{\min}
+
(w_{\max}-w_{\min})\phi(z_i),
\label{eq:adaptive_weight}
\end{equation}
where \(\phi:[0,1]\rightarrow[0,1]\) is a monotonic mapping. Larger model discrepancy therefore results in a larger weight and reduces reliance on the corresponding section.

Although \(W\) may change between control steps, it remains fixed throughout the projection iterations within each step. Each filtering problem therefore retains the fixed-metric projection structure described above. Algorithm~\ref{alg:wcbf} summarizes the resulting control loop.


The weight matrix $W$ remains fixed during the projection iterations within each control step, ensuring convergence \cite{bregman1965method}. Weight updates between steps do not change the CBF constraints or the associated invariance conditions.

\begin{algorithm}[t]
\caption{Adaptive W-CBF Control Loop}
\label{alg:wcbf}
\begin{algorithmic}[1]
\REQUIRE barriers $h_1,\ldots,h_M$ with function $\alpha$, weight mapping $\phi$, nominal command $u_{\mathrm{nom}}$
\FOR{every control step}
\STATE measure the rack lengths $x$ and the frame poses
\IF{the update window $\mathcal B_k$ is complete}
\STATE update $\eta_i,z_i,w_i$ by \eqref{eq:adaptive_indicator}--\eqref{eq:adaptive_weight} and form $W$ by \eqref{eq:weight_matrix}
\ENDIF
\STATE $a_i=\nabla h_i(x)$, $b_i=-\alpha(h_i(x))$, $i=1,\ldots,M$
\STATE $u=u_{\mathrm{nom}}$
\FOR{a fixed number of sweeps}
\FOR{$i=1$ to $M$}
\STATE project $u$ onto $\mathcal H_i$ by \eqref{eq:wcbf_closed_form}
\ENDFOR
\ENDFOR
\STATE evaluate the residual $r(u)$ by \eqref{eq:projection_residual}
\STATE apply $u$ to the racks
\ENDFOR
\end{algorithmic}
\end{algorithm}




\section{Experimental Setup and Results}
To evaluate the performance of the proposed controller, hardware experiments were conducted on the robotic platform described in Sec. II. During the experiments, the motors were operated in speed mode and limited to 36 RPM. The control loop was operated at 25~Hz while receiving pose information from the Lighthouse positioning system at 30~Hz.

Three types of constraints are imposed on the robot: collision-avoidance constraints, rack-bending constraints, and actuation constraints. For collision detection, each robot segment is modeled as an oriented rectangle, while the flexible racks are approximated by transverse capsule slices distributed along their centerlines. This representation enables whole-body collision checking with respect to the surrounding environment. A minimum safety margin of \(20~\mathrm{mm}\) is enforced.

The rack-bending constraints prevent excessive curvature by limiting the relative deformation between the two sides of each segment. The actuation constraints restrict each rack displacement to \(10\text{--}200~\mathrm{mm}\), the rack speed to \(30~\mathrm{mm/s}\), and the tip speed to \(120~\mathrm{mm/s}\). Although the input limits are not formulated as control barrier functions, they can be expressed as half-space constraints in the input space and handled using the same projection formulation.

The three types of constraints are treated with equal importance, and we assume that their feasible sets have a nonempty intersection. For additional protection of the robot, the collision-avoidance constraints are projected last, such that the resulting collision-safe command is not further modified by subsequent projections.

\begin{figure*}[!t]
    \centering
    \includegraphics[width=0.8\textwidth]{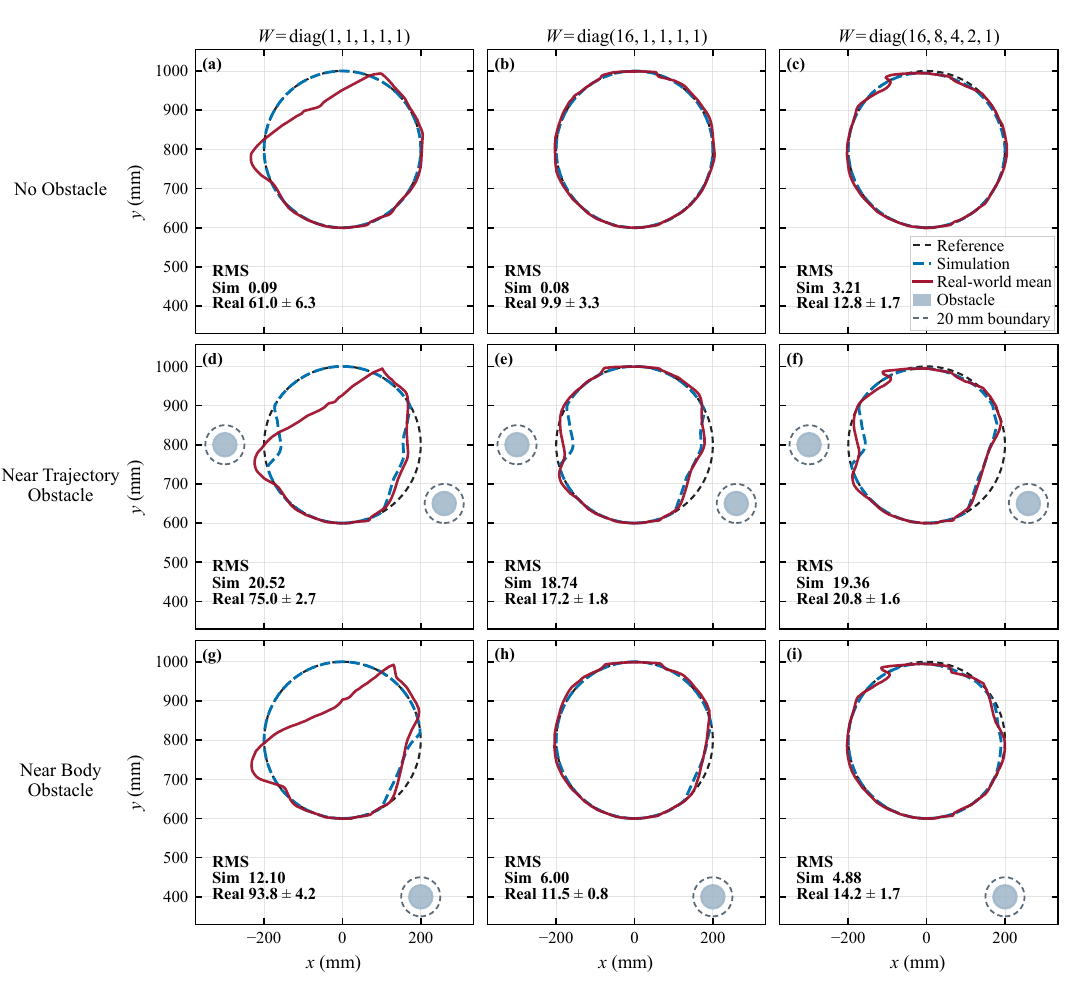}
    \caption{Results of the W-CBF controller following a circular trajectory in simulation and on the hardware, with the three scenarios by row and the three weight matrices by column. The curves show the path of the tip center, the hardware curve being the phase average of three runs.}
    \label{fig3}
    \vspace{-0.2in}
\end{figure*}

\begin{figure}[!b]
    \centering
    \includegraphics[width=0.45\textwidth]{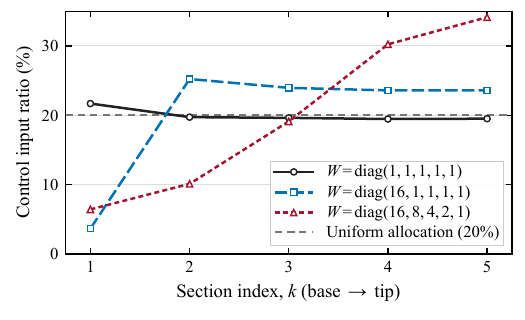}
    \caption{Control input share of each segment under the three fixed weight matrices, across all scenarios and runs.}
    \label{fig4}
\end{figure}

In the experiments, we used 10 projection sweeps per step, with $\alpha(h)=h$ for obstacle constraints and $\alpha(h)=2h$ for rack coupling and actuation constraints. The update window $\mathcal{B}_k$ was nominally 0.1~s, with $(\eta_{\min},\eta_{\max})=(0.05,0.25)$, $(w_{\min},w_{\max})=(1,16)$, $L_0=30~\mathrm{mm}$ and $k_s=10$. Since the cyclic projections are closed-form, the safety filter required less than 0.6~ms per control step in all three scenarios ($0.534\pm0.087$, $0.575\pm0.090$ and $0.542\pm0.112$~ms). For the cleaning task, since the route becomes complex and the constraints become stricter, we raised the projection budget to 500 sweeps per step, guaranteeing convergence, and shortened the obstacle $\alpha(h)$ to $0.5h$, slowing down the approach, resulting in a solver time of $10.339\pm0.500$~ms.


\begin{figure*}[!t]
    \centering
    \includegraphics[width=.8\textwidth]{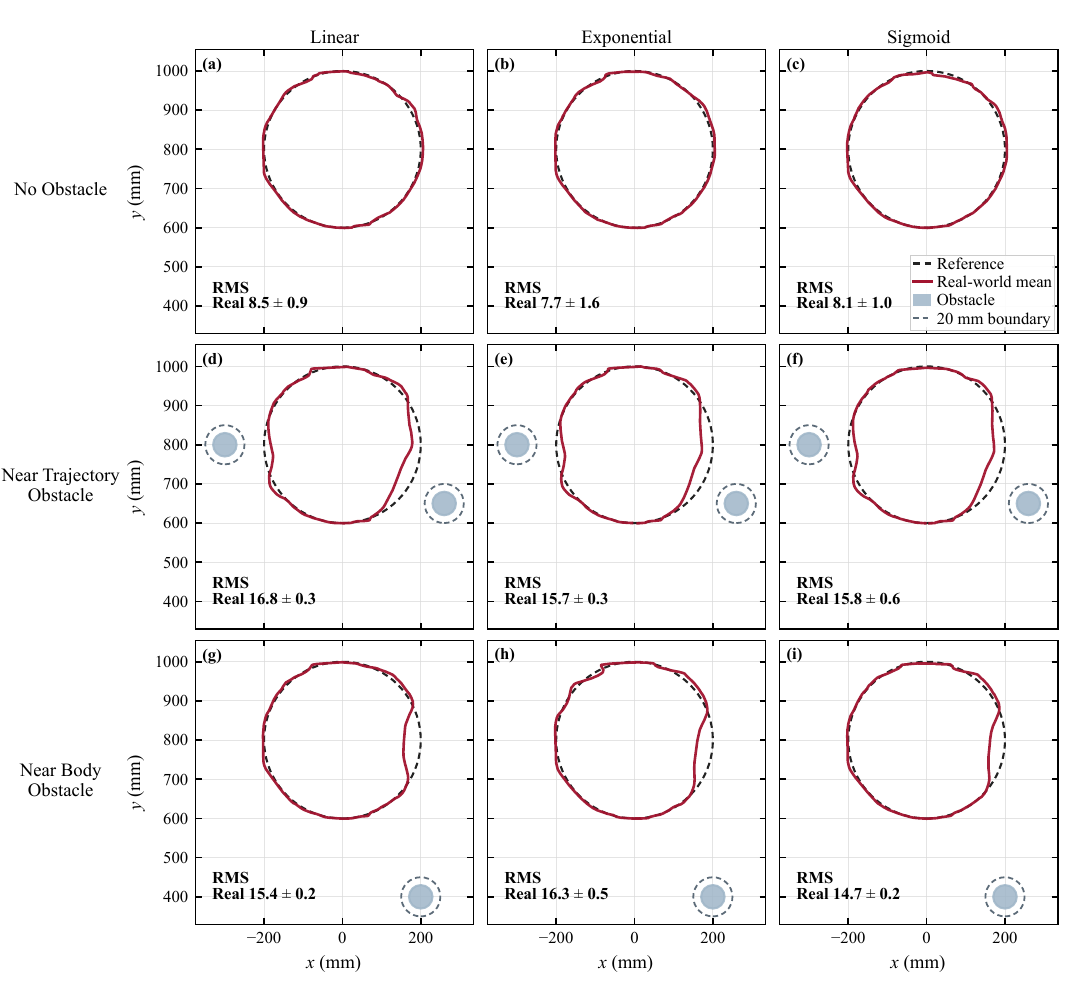}
    \caption{Results of the adaptive W-CBF controller following a circular trajectory on the hardware, with the three scenarios by row and the linear, exponential and sigmoid mappings by column. The curves show the path of the tip center, averaged in phase over three runs.}
    \label{fig5}
    \vspace{-0.2in}
\end{figure*}

\subsection{W-CBF Controller}\label{W-CBF}
A circular trajectory tracking task was first employed to test the W-CBF controller. The reference circle had a radius of 200~mm, was centered at $(0,800)$~mm in the robot base frame, and the tip traversed it at 30~mm/s. To investigate the influence of obstacles, we designed three scenarios: no obstacles, two obstacles placed 100~mm outside the reference circle on opposite sides of the trajectory, and a single obstacle beside the robot body. Each obstacle had a radius of 30~mm. In each scenario, three weight matrices were tested, with the form of Eq. \eqref{eq:weight_matrix} ensuring two racks of each segment share the same weight. From the base to the tip, the segment weights are $[1,1,1,1,1]$, $[16,1,1,1,1]$ and $[16,8,4,2,1]$, denoted by $W_1, W_2$ and $W_3$, representing the uniform baseline, increasing weighting of only the proximal segment, and gradually weighting along the robot. Identical controller parameters are employed in the simulation and physical experiment. 

\begin{table}[!b]
\renewcommand{\arraystretch}{1.1}
\setlength{\tabcolsep}{2pt}

\caption{Tip Tracking RMS Error in Simulation (Sim.) and on Hardware (HW)
Under Weight Matrices (RMS: mm; $\Delta$: \%)}
\label{tab:rms}
\centering
\begin{tabular}{lccc@{\hspace{6pt}}ccc@{\hspace{6pt}}ccc}
\hline\hline
 & \multicolumn{3}{c}{No obstacle} & \multicolumn{3}{c}{Near-trajectory} & \multicolumn{3}{c}{Near-body} \\
\cline{2-4}\cline{5-7}\cline{8-10}
 & Sim. & HW & $\Delta$ & Sim. & HW & $\Delta$  & Sim. & HW & $\Delta$  \\
\hline
$W_1$ & 0.09 & 61.0{\tiny$\pm$6.3} & --  & 20.52 & 75.0{\tiny$\pm$2.7} & --  & 12.10 & 93.8{\tiny$\pm$4.2} & -- \\
$W_2$ & 0.08 & 9.9{\tiny$\pm$3.3} & 84  & 18.74 & 17.2{\tiny$\pm$1.8} & 77  & 6.00 & 11.5{\tiny$\pm$0.8} & 88\\
$W_3$ & 3.21 & 12.8{\tiny$\pm$1.7} & 79 & 19.36 & 20.8{\tiny$\pm$1.6} & 72  & 4.88 & 14.2{\tiny$\pm$1.7} & 85\\
\hline\hline
\end{tabular}

\vspace{2pt}
\begin{minipage}{\columnwidth}\footnotesize
$\Delta$ is the reduction of the hardware RMS relative to $W_1$ in
the same scenario.
\end{minipage}
\vspace{-0.3in}
\end{table}  

Fig.~\ref{fig3} and Table~\ref{tab:rms} show that the weighting has limited effect in simulation but is decisive on the hardware. Under the uniform baseline $W_1$, the simulated tip follows the reference to 0.09~mm RMS without obstacles, while the real robot fails to track the quarter circle in the upper-left corner. Similar deviations occur in both obstacle scenarios, with RMS errors of 75.0 and 93.8~mm. Under uniform weights, each section carries about one fifth of the commanded rack motion (Fig.~\ref{fig4}). However, motion of the proximal sections must overcome friction accumulated along the downstream chain, leading to localized tracking errors. In contrast, the simulation neglects friction and rack deflection, resulting in an almost ideal circular trajectory under the same weighting.

Both nonuniform weight matrices redistribute the control input by taking the motion away from the proximal part, whose share falls to 3.7\% under $W_2$ and 6.4\% under $W_3$ (Fig.~\ref{fig4}). Without obstacles, the hardware error drops from 61.0~mm to 9.9~mm under $W_2$, while the simulated error is unchanged, indicating the weight matrices work against the discrepancies. For obstacles located near the reference trajectory, the hardware tracking error under $W_2$ was 17.2~mm, slightly lower than the simulated value of 18.7~mm. However, the reduced RMS error should be interpreted together with the observed safety-margin incursion in the physical experiments. While the robot remained close to the safety boundary in simulation, small deviations were observed experimentally due to kinematic discrepancies. Importantly, nonuniform weighting consistently reduced the maximum estimated incursion from 40.78~mm with $W_1$ to 25.75~mm with $W_2$ and 20.43~mm with $W_3$, demonstrating improved robustness of the safety behavior under physical implementation.

\subsection{Adaptive W-CBF Controller}\label{AW-CBF}
Although fixed weights perform well across the tested tasks, no single configuration is optimal in all cases, partly because model discrepancies vary with the robot state. We therefore adapt the weights using a segment-wise indicator of model discrepancy and evaluate three mappings from this indicator to the corresponding weights.

Different mapping functions produce different segment weights for the same model discrepancy, thereby affecting motion allocation. We consider linear, exponential, and sigmoid mappings with the same weight bounds. The linear mapping increases the weight at a constant rate. The exponential mapping responds more strongly as the error approaches the upper threshold, while the sigmoid mapping emphasizes changes around the midpoint and is less sensitive near either bound. We evaluate their effects on tracking performance under the same trajectory and obstacle conditions used in the fixed weight experiments. All three functions follow Eq. \eqref{eq:adaptive_weight}, with the shape of the interpolation differing as:

\begin{equation*}
\begin{aligned}
\phi_{\mathrm{lin}}(z)&=z,\qquad
\phi_{\mathrm{exp}}(z)=\frac{r^{z}-1}{r-1},\\[2pt]
\phi_{\mathrm{sig}}(z)&=\frac{\sigma\!\left(k_s(z-\tfrac12)\right)-a}{1-2a},
\end{aligned}
\end{equation*}

where $r=w_{\max}/w_{\min}$, $\sigma(x)=1/(1+e^{-x})$ and $a=\sigma(-k_s/2)$. Larger weights are assigned to segments with greater error, reducing reliance on their actuation. This construction also explains why the adaptive controller is evaluated on the hardware alone. The simulator integrates the same kinematic model that the controller uses to predict, so the residuals in Eq. \eqref{eq:adaptive_indicator} vanish identically, and the adaptive controller degenerates to the uniform baseline that was already reported in Fig.~\ref{fig3}. 

Fig.~\ref{fig5} compares the hardware trajectories under adaptive weighting. With near-trajectory obstacles, all three adaptive mapping functions achieve lower RMS error compared with 17.2~mm for $W_2$. With a near-body obstacle, the adaptive mappings produce higher errors than the two nonuniform fixed settings, with the best adaptive result reaching 14.7~mm compared with 11.5~mm for $W_2$. This difference may arise because the adaptive indicator measures local model discrepancy rather than the contribution of proximal motion to whole-body obstacle avoidance. It may therefore reduce the proximal weight when the discrepancy is small, whereas $W_2$ maintains the same penalty throughout the task.

\begin{figure}[!t]
    \centering
    \includegraphics[width=0.45\textwidth]{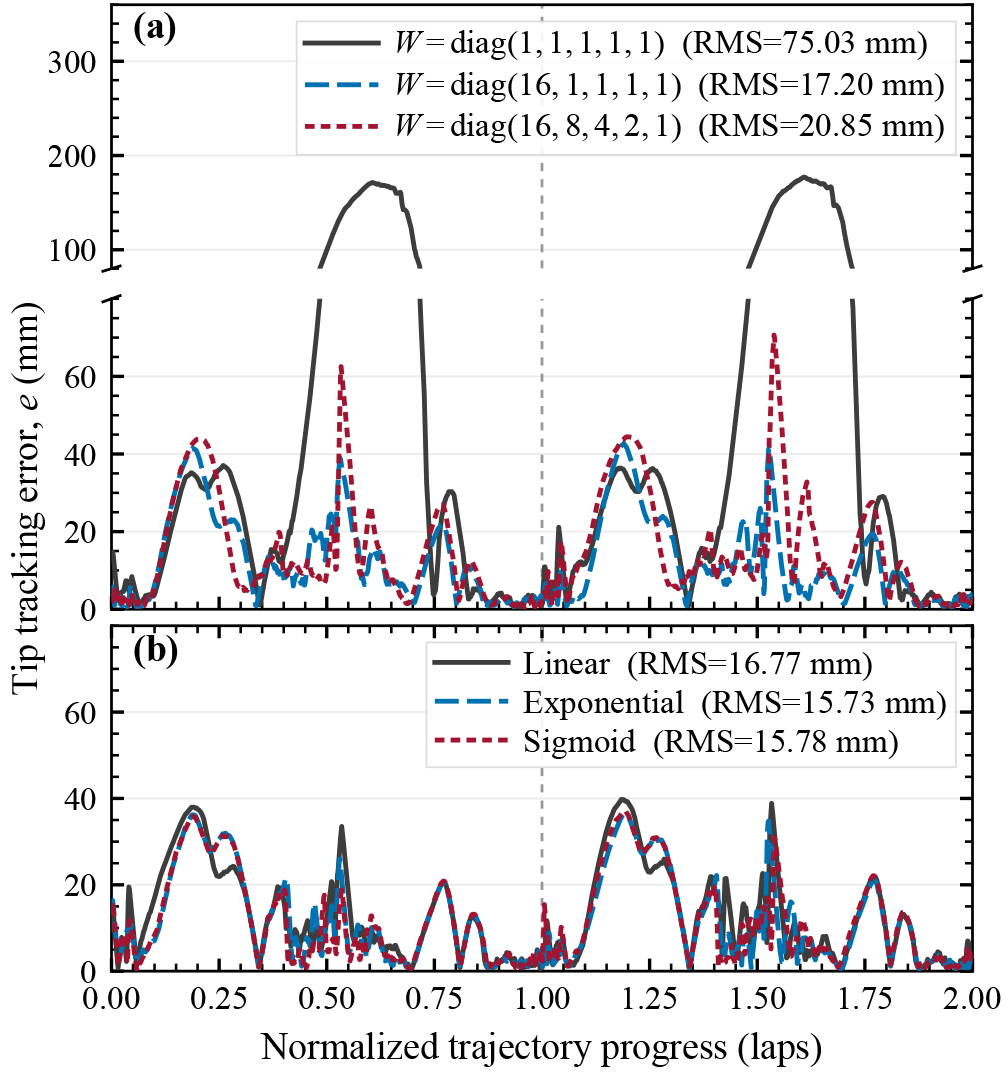}
    \caption{Tip tracking error over two laps with obstacles near the trajectory. The lap starts at the bottom of the circle and runs counterclockwise. (a) The three fixed weight matrices. (b) The three adaptive weight matrices.}
    \label{fig6}
    \vspace{-0.2in}
\end{figure}

Fig.~\ref{fig6} compares the tracking error over two laps with obstacles near the reference path.
The two laps are almost identical, with each setting peaking at the same points in both. The largest error occurs in the upper left corner, where the uniform baseline holds a broad plateau near 150~mm. Weighting removes the plateau. The median error over that quarter drops below 10~mm, and only a narrow spike of about 41~mm remains. $W_3$ leaves a larger one, 71~mm. The trials with the right obstacle are different. Its peak is near 36~mm under uniform weights. The persistent peak near the right obstacle is consistent with the deviation required by obstacle avoidance. Weighting therefore acts mainly on the part of the error that the model gets wrong, and it removes most of it. 

\begin{figure}[t]
    \centering
    \includegraphics[width=\columnwidth]{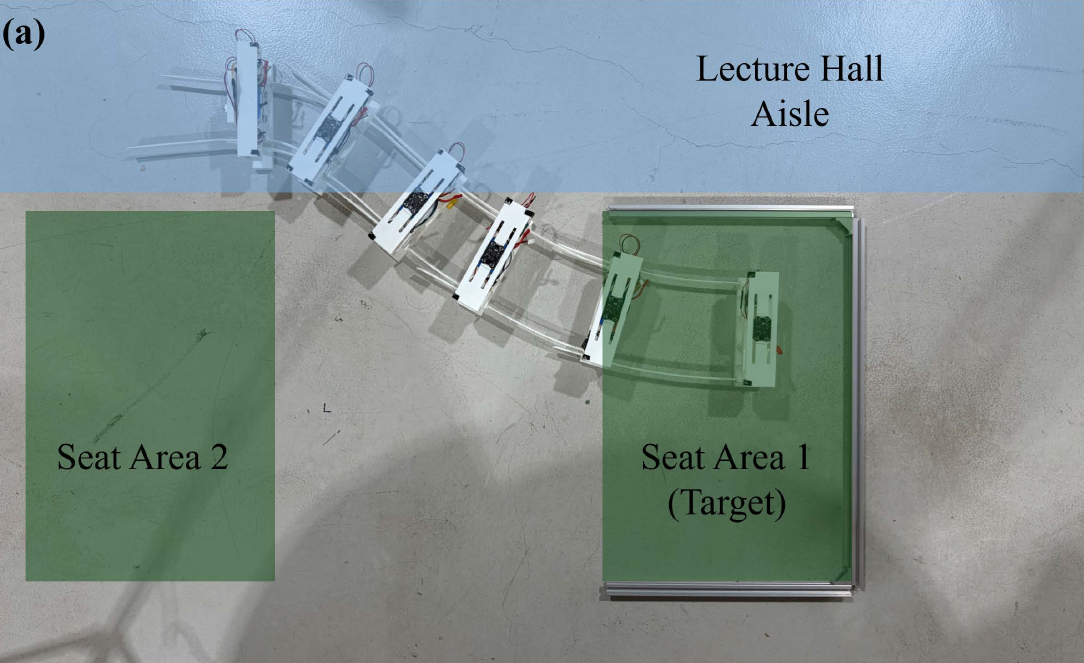}\\[5pt]
    \includegraphics[width=\columnwidth]{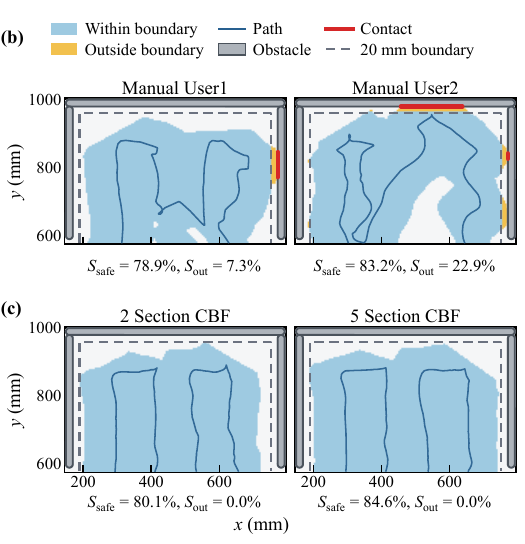}
    \caption{Real-world cleaning task and end frame coverage.
      (a)~Experimental setup.
      (b)~Manual joystick teleoperation by two operators.
      (c)~Adaptive W-CBF control with 2 equivalent sections and 5 sections.
      $S_\mathrm{safe}$ is the covered fraction of the target area
      ($213.70\times10^{3}$\,mm$^2$); $S_\mathrm{out}$ is the covered fraction
      of the 20\,mm safety band ($27.25\times10^{3}$\,mm$^2$).}
    \label{fig:cleaning_coverage}
    \vspace{-0.2in}
\end{figure}

\subsection{Cleaning Tasks}\label{Cleaning}
The last experiment evaluated the controller in a cleaning task that modeled cleaning between the rows of a lecture hall. The surrounding seats form a narrow corridor containing the target cleaning area. Fig.~\ref{fig:cleaning_coverage}(a) shows the setup, where barriers restrict the robot's motion and the end frame sweeps the designated region.

We first recorded a manual baseline. Two operators were asked to control the robot using a joystick. Limited by the interfaces, the robot was controlled under only two equivalent segments: the two proximal segments and the three distal ones. Accordingly, we ran adaptive W-CBF control twice, including 2 equivalent segments, for comparison with the operators, and the 5 segments, with each segment driven independently and all ten degrees of freedom available. The reference sweep path consists of four equally spaced parallel passes with a spacing of 121~mm and a length of 329~mm. Adjacent passes are connected by lateral moves at alternating ends. The path accounts for the end-frame size, panel radius, and safety margin. The tip follows the reference at 30~mm/s, while the CBF filter handles whole-body constraints. Both used sigmoid weighting, and each trial ended after two complete up and down sweeps. Coverage was evaluated from the end-frame swept area reconstructed using Lighthouse. 

Fig.~\ref{fig:cleaning_coverage}(b) shows the coverage achieved by the two operators, reflecting a conservative and an aggressive strategy, respectively. The aggressive operator swept 83.2\% of the target area but also covered 22.9\% of the safety band. The conservative one limited safety-band coverage to 7.3\% but achieved only 78.9\% coverage of the target area. Neither avoided the band, and both reached the barrier surface itself, over a length of 74~mm for one operator and 196~mm for the other.

By contrast, Fig.~\ref{fig:cleaning_coverage}(c) shows the two adaptive W-CBF runs. Both kept the swept area out of the safety band entirely and never reached a barrier. With the same configurations as the manual operation, the controller covered 80.1\% of the target area, which sits between the two manual trials. Regarding five equivalent sections, it reached 84.6\%, above both of them. It shows that the extra DoFs of the hyper-redundant robot contribute to performance under conditions with complex obstacles. Together, these results show that adaptive W-CBF control expands coverage while keeping the measured sweep inside the safety boundaries.

\section{Conclusion and Future Work}
In this work, we proposed a weighted control barrier function (W-CBF) controller for safety control of hyper-redundant robots, together with an investigation of adaptive methods that tune the weights online from measured model error. The weight matrix only redistributes the control action among the sections, while remaining compatible with the CBF constraints. In physical experiments, this choice proved almost inert in simulation but decisive on hardware, reducing the circular tracking error from 61.0 to 9.9~mm. Fixed weights require offline tuning, and suboptimal choices may degrade tracking accuracy. In contrast, the adaptive method achieves most of the performance improvement without requiring manual selection of segment-specific weights. In a constrained cleaning task, the controller kept the swept end frame clear of the safety band while covering 84.6\% of the region, matching or exceeding the best human-operated performance without the frequent collisions observed during manual operation. Notably, the robot was not specifically optimized to achieve maximum coverage, but moves its end-effector along a designated path. 

Although the adaptive weighted CBF controller performs well in the current experiments, it is difficult to conclude that the present weighting strategy is optimal. Further investigation is needed, particularly for the normalized discrepancy mapping method. In addition, safety margin violations were observed during the experiments, potentially due to model mismatch arising from friction, inertial effects, and rack deformation. Future work will investigate adaptive dynamic models that capture these effects and their variations over time, potentially using neural networks to improve motion prediction and constraint satisfaction.

\bibliographystyle{IEEEtran}
\bibliography{reference}

\end{document}